\documentclass[a4paper]{ifacconf}

\counterwithin*{section}{part}
\usepackage{cite}
\usepackage{amsmath}
\usepackage{amssymb}
\usepackage{graphicx}
\usepackage{textcomp}
\usepackage{xcolor}
\usepackage{pgfplots}
\usepackage{subfigure}
\usepgfplotslibrary{groupplots}
\usepackage{tikz}
\usepackage{tikzscale}
\usepackage{cleveref}
\usepackage{float}
\usepackage{multirow}
\usepackage{algorithm}
\usepackage{algpseudocode} 
\usepackage{siunitx}
\usepackage{comment}
\usepackage{booktabs}
\usepackage[square,numbers]{natbib}
\def\BibTeX{{\rm B\kern-.05em{\sc i\kern-.025em b}\kern-.08em
    T\kern-.1667em\lower.7ex\hbox{E}\kern-.125emX}}
    
\crefname{figure}{Fig.}{Fig.}
\Crefname{figure}{Figure}{Figures}
\crefname{equation}{}{}
\Crefname{equation}{Equation}{Equations}

\newcommand{\ie}{\textit{i}.\textit{e}., }
\newcommand{\eg}{\textit{e}.\textit{g}., }
\newcommand{\st}{\text{s.t. }}
\DeclareMathOperator*{\argmin}{arg\,min}

\begin{document}
\begin{frontmatter}

\title{Task-Agnostic Adaptation for \\ Safe Human-Robot Handover} 

\thanks[footnoteinfo]{This work is in part supported by Siemens. The robot arm is donated by FANUC Corporation.}

\author[First]{Ruixuan Liu} 
\author[First]{Rui Chen} 
\author[First]{Changliu Liu}

\address[First]{Robotics Institute,
	Carnegie Mellon University,
	Pittsburgh, PA, 15213, USA (e-mail: ruixuanl, ruic3, cliu6@andrew.cmu.edu).}

\begin{abstract}      
Human-robot interaction (HRI) is an important component to improve the flexibility of modern production lines.
However, in real-world applications, the task (\ie the conditions that the robot needs to operate on, such as the environmental lighting condition, the human subjects to interact with, and the hardware platforms) may vary and it remains challenging to optimally and efficiently configure and adapt the robotic system under these changing tasks.
To address the challenge, this paper proposes a task-agnostic adaptable controller that can 1) adapt to different lighting conditions, 2) adapt to individual behaviors and ensure safety when interacting with different humans, and 3) enable easy transfer across robot platforms with different control interfaces.
The proposed framework is tested on a human-robot handover task using the FANUC LR Mate 200id/7L robot and the Kinova Gen3 robot.
Experiments show that the proposed task-agnostic controller can achieve consistent performance across different tasks.
\end{abstract}


\end{frontmatter}

\section{Introduction}
\begin{figure}
\subfigure[]{\includegraphics[width=0.45\linewidth]{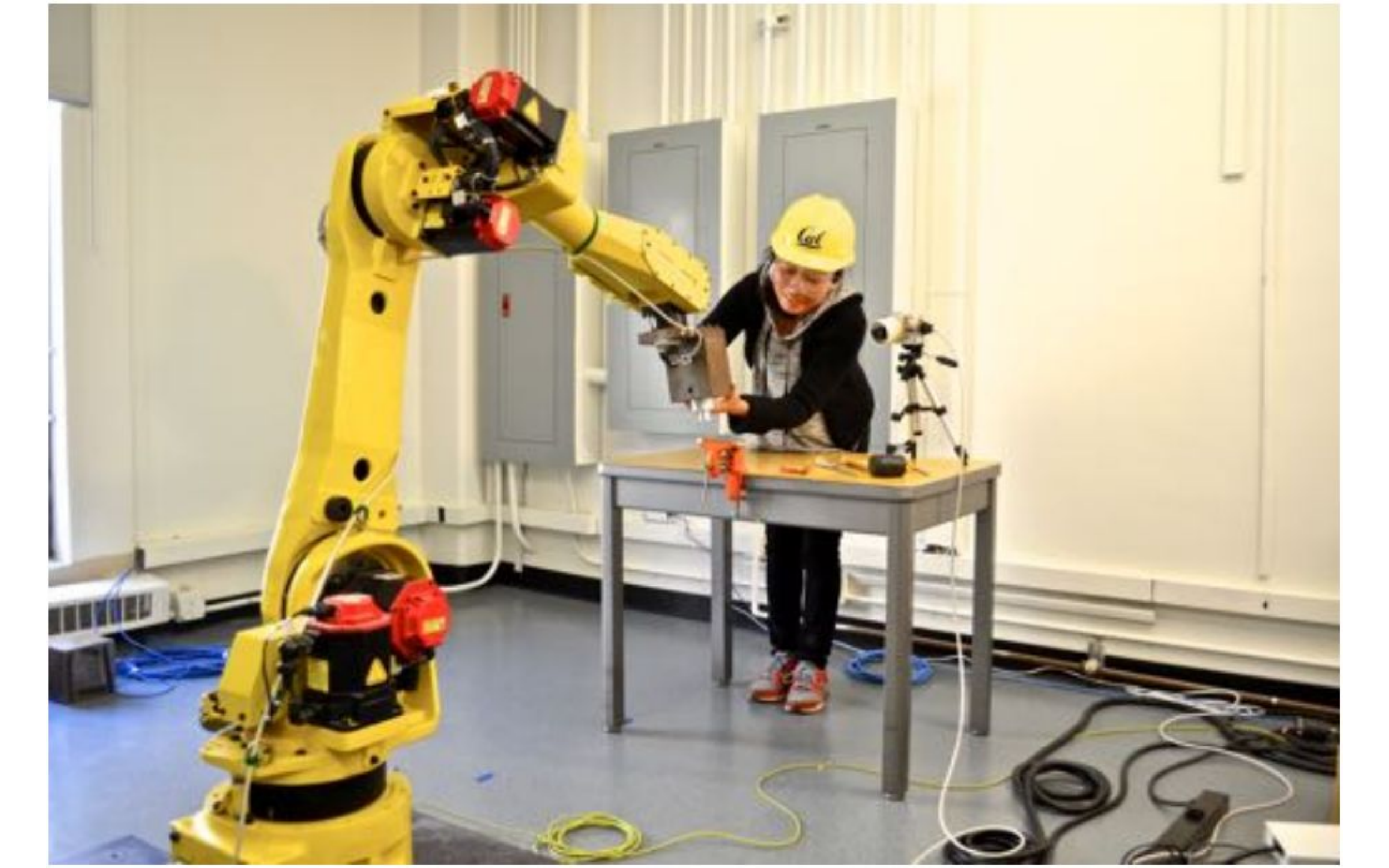}\label{fig:hri1}}\hfill
\subfigure[]{\includegraphics[width=0.45\linewidth]{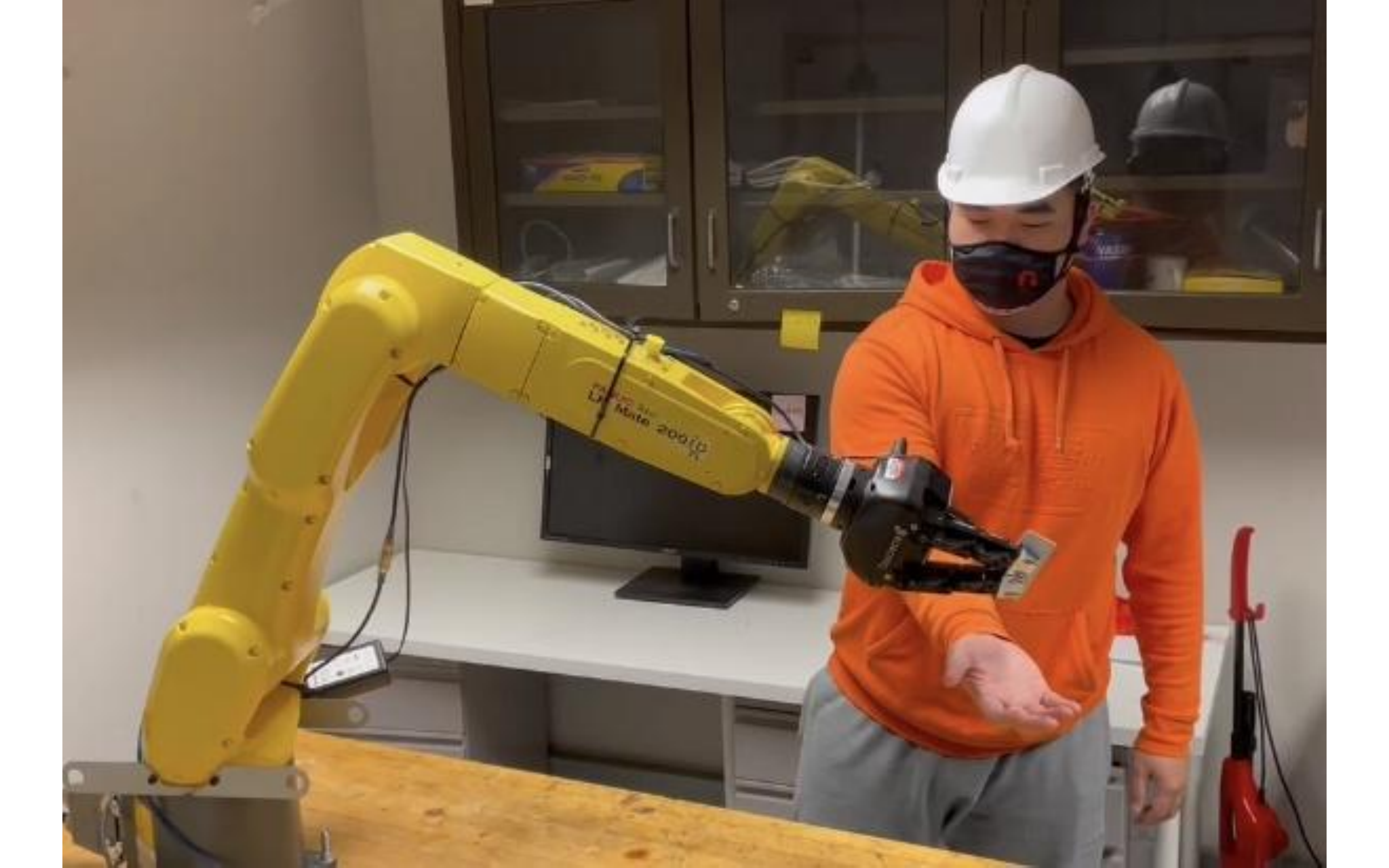}\label{fig:hri2}}
\vspace{-10pt}
    \caption{\footnotesize Examples of HRI. (a) Human-robot co-assembly \cite{7487476}. (b) Human-robot handover \cite{jssa}. \label{fig:hri}}
    \vspace{-10pt}
\end{figure}

Industrial robots are widely used in many applications \cite{GOPINATH2017430, 6840175} with structured and static environments.
However, the need to increase the flexibility of production lines is calling for robots to work in dynamic environments, such as collaborating with human workers (HRI) \cite{Christensen2021ARF, KRUGER2009628}.
\Cref{fig:hri} shows examples of HRI, where the robot co-assembles an object with the human in \cref{fig:hri1} and the robot delivers a desired workpiece to the human in \cref{fig:hri2}.  

The current industrial practice is to manually tune the whole system case by case, \ie to specific tasks and hardware platforms in different applications.
Such manual tuning is only practical for tasks that are fully specified and run for long periods of time without any change.
In typical HRI settings, however, the task may often vary in unpredictable ways and it is expensive, if not impossible, to design a task-specific robotic system for every single scenario.
For example, different human workers might interact with the robot in personalized ways which are difficult to anticipate beforehand.
Hence, to reduce the development cost, we desire to adapt existing robotic systems to different tasks with minimal tuning.

More formally, this paper focuses on task-agnostic adaptation for HRI. We consider adaptation in three major aspects of HRI tasks: \textit{objective} (\eg the target robot configuration), \textit{constraint} (\eg the collision avoidance constraint), and \textit{execution} (\eg the robot actual motion profile during HRI, including speed, acceleration, etc).
We desire the adaptation to be task-agnostic such that the same techniques can be directly applied to various HRI tasks. For clarity, the rest of the discussion in this paper focuses on the task of human-robot handover (\cref{fig:hri2}), but note that the same adaptation methods can be directly applied to other HRI tasks.
In general, human-robot handover is a class of tasks that requires the robot to deliver a human-desired object.
A typical process of such kind follows the order of perception (\ie to determine the \textbf{objective} from the perception of the environment and the human),
motion planning (\ie to enforce the \textbf{constraint} of safe and efficient robot motions), and control (\ie to command the \textbf{execution} of planned trajectories).
All of these steps need adaptation when tasks vary: the perception might be inaccurate under poor lighting conditions; the comfortable robot handover pose might be different for different humans; the execution might require different robot control interfaces. Therefore, it is a challenge to achieve adaptable HRI tasks.

To address the challenge, this paper develops a controller that solves the task-specific objective while satisfying both hardware and safety constraints and adapts to different HRI tasks with minimum tuning. The controller can 1) adapt the system to different environment lighting conditions, 2) adjust the system behavior according to different human subjects, and 3) enable the transferability across robot platforms with minimum tuning effort. 
To adapt to the lighting conditions, we adjust the delivery motion according to the perceptual uncertainty. To adapt to different humans, we optimize the delivery pose based on human behavior. To minimize the transfer effort, we divide the control software into a task-level controller and a motion-level controller. 
To demonstrate the system, we perform the human-robot handover task on both the FANUC LR Mate 200id/7L and the Kinova Gen3 robots.   
Experiments show that the proposed framework can adapt and enable similar safe interaction with changing tasks.

The following paper is organized as follows. 
\Cref{sec:so} introduces the hardware in use. 
\Cref{sec:method} proposes the task-agnostic adaptation. 
Experiment results are shown in \cref{sec:exp}. 
\Cref{sec:conclusion} concludes the paper.

\section{Hardware Overview}\label{sec:so}
In this section, we introduce the hardware used in this paper. 
To demonstrate the task-agnostic system, we perform the human-robot handover task on a FANUC LR Mate 200id/7L robot arm and a Kinova Gen3 robot arm.   

\begin{figure}
\footnotesize
    \centering
    \includegraphics[width=0.9\linewidth]{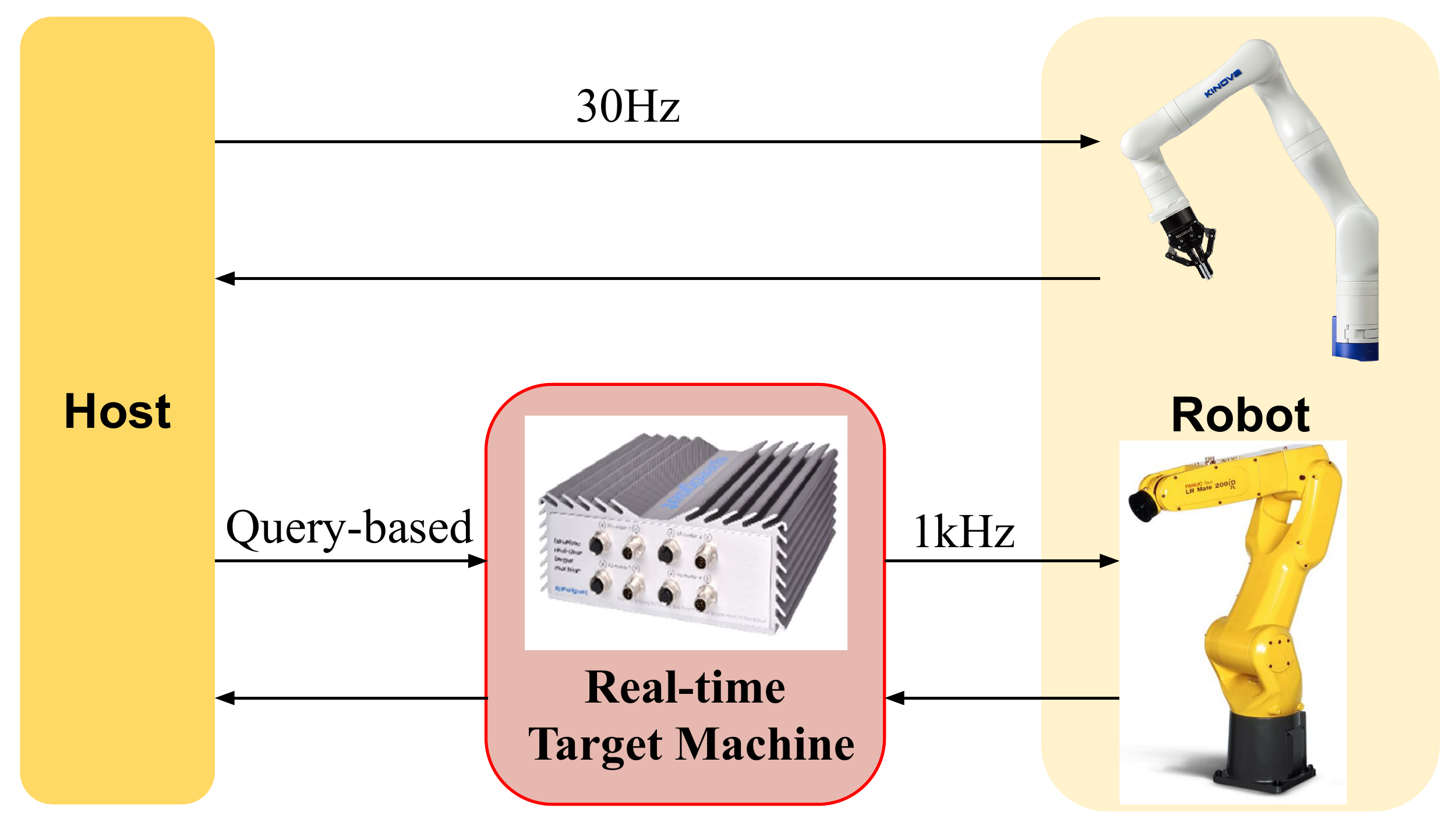}
    \caption{\footnotesize Hardware connection diagram.}
    \label{fig:hardware_connection}
\end{figure}

\subsection{FANUC LR Mate 200id/7L}
The FANUC LR Mate 200id/7L robot is shown in the bottom-right of \cref{fig:hardware_connection}. 
It is an industrial robot widely used in fields that require high precision.
It provides a position control interface via Ethernet (called \textit{stream motion}), which requires a 125 Hz position control sequence with bounded jerks. 
In addition, according to our practical experience, the position control interface requires a stable 1 kHz communication for stable performance. 
It is difficult, if not impossible, to maintain such a high-frequency stable communication using a general-purpose PC.
Therefore, we have a Speedgoat baseline real-time target machine (SG) bridging the robot and the host as shown in \cref{fig:hardware_connection}. 
The host and the SG have a query-based Ethernet communication depending on the task status, and the SG communicates with the robot at 1 kHz.

\subsection{Kinova Gen3}
The Kinova Gen3 robot is shown in the top-right of \cref{fig:hardware_connection}. 
It is a lightweight collaborative robot widely used in HRI. 
Unlike the FANUC robot, the Kinova Gen3 does not require high-frequency communication, and thus, the host directly controls the robot via Ethernet as shown in \cref{fig:hardware_connection}. 
The Kinova robot provides the Kinova Kortex API\footnote{https://github.com/Kinovarobotics/kortex}, a controller interface in C++.
The host and the robot have communication at approximately 30 Hz on average.

\section{Task-agnostic Adaptation}\label{sec:method}

A typical HRI task can be formulated as constrained optimization. The objective is to find an optimal robot trajectory, while the constraints are enforced to meet safety and hardware requirements. Let $q=[q_1;q_2;\dots;q_n]$ denote the joint trajectories of length $n$ and $E$ denote the environment (\ie humans and obstacles poses), we can write the general formulation of HRI tasks as
\begin{equation}\label{eq:general_formulation} 
    \argmin_q ~ L(q, E) ~
    \st ~ C_\mathrm{hardware}(q), ~C_\mathrm{safety}(q, E), 
\end{equation}
where $L$ represents the task-specific objective, $C_\mathrm{hardware}$ encodes hardware-related constraints such as dynamics limit and kinematics constraints, and $C_\mathrm{safety}$ guarantees safety.
To solve \eqref{eq:general_formulation} with adaptability to various tasks, we decompose the optimization into several modules.
With such a design, the variability of each task attribute is handled explicitly by a certain module with others unchanged.

The first module is a function $g(\cdot)$ that determines the end-effector Cartesian goal pose $x_G$ determined by visual perception, \ie $x_G = g(E)$. 
HRI tasks with multiple stages can be handled simply by invoking $g(\cdot)$ each time a new goal is desired.
For instance, human-robot handover includes several stages, \ie 1) idle, 2) reach the target object, 3) move to the deliver pose, 4) return the tool if needed, 5) return to home.
The second module is a function $f(\cdot)$ that calculates the desired robot joint configuration $q_g$ given the environment and the end-effector Cartesian goal, \ie $q_g=f(x_G, E)$. 
The function $f(\cdot)$ introduces extra freedom in adapting the robot motion to factors beyond task objectives. For example, the robot needs to keep a safe distance between the human and any of its links. 
At the same time, the $q_g$ should always guarantee the correct end-effector pose $x_G$ according to robot-specific forward kinematics. This constraint is included in $C_\mathrm{hardware}$ and $C_\mathrm{safety}$ in \eqref{eq:general_formulation}.
The third module is a tracking controller that tracks $q_g$ and minimizes $L$ in the long term. This module handles the adaptation to different hardware platforms with different dynamics (\eg acceleration or jerk) constraints and programming interfaces, and is included in $C_\mathrm{hardware}$ in \eqref{eq:general_formulation}.
The final module is a safe controller that ensures short-term safety constraints (\eg collision avoidance) by actively monitoring the minimum distance between all robot links and the human. 
This corresponds to $C_\mathrm{safety}$ in \eqref{eq:general_formulation}.

With the above decomposition, \eqref{eq:general_formulation} can be re-written as 
\begin{equation}\label{eq:prob}
    \begin{split}
        \argmin_q~&~L_T(q)\\
        \st~&~\underbrace{x_G=\mathbb{E}\left[g(E_n)\right],~q_g=q_n=\mathbb{E}\left[f(x_G, E_n)\right]}_{C_\mathrm{hardware}},\\
        &~\underbrace{\forall k\in[1,n], ~\mathbb{E}\left[d(q_k,E_k)\right]\geq d_{min}}_{C_\mathrm{safety}},
    \end{split}
\end{equation}
for a single stage in an HRI task,
where $L_T(\cdot)$ is a tracking cost function and $d_{min}$ is the safety margin. $E_n, E_k$ are the environments at the final and intermediate timesteps, which are obtained using perception algorithms, \eg OpenPose \cite{openpose}.
And $d(\cdot)$ is a distance function that calculates the minimum distance between the robot links and the environment objects using the capsule representation \cite{7487476}. 
Notably, we consider \eqref{eq:prob} in a probabilistic sense due to the uncertainty in human motions. Hence, we take expectations of the quantities in constraints over possible environment states.
The solved joint trajectory $q$ is sent to the hardware via appropriate control interfaces for execution. During the task, the four modules run at different frequencies. Both $g(\cdot)$ and $f(\cdot)$ are invoked on demand per new tracking goal, while the tracking controller keeps running throughout the task to solve \eqref{eq:prob} repeatedly.
The safe controller enforces $C_\mathrm{safety}$ at a higher frequency than the tracking controller, similar to \cite{serocs}. In this way, safety can be assured even when the other modules need a nontrivial amount of time to respond to environmental changes.

Existing motion planners \cite{cfs,dscc_cfs} can solve the joint trajectory $q$ from the optimization in \cref{eq:prob}. 
However, the formulation is prone to task change, \ie $g(\cdot)$ and $f(\cdot)$ might be different.
To achieve task-agnostic safe interaction, we need to ensure that 1) the mapping $g(\cdot)$ can calculate the proper $x_G$ according to \textit{different lighting conditions}, 2) the mapping $f(\cdot)$ can compute the appropriate $q_g$ when interacting with \textit{different humans} and 3) the control architecture can easily transfer between \textit{different robots} to execute $q$. In the rest of this section, we elaborate on each of these adaptation aspects.

\subsection{Adaptation to Different Lighting Conditions}

\begin{figure}
\subfigure[\footnotesize Bright: deliver to human hand.]{\includegraphics[width=0.45\linewidth]{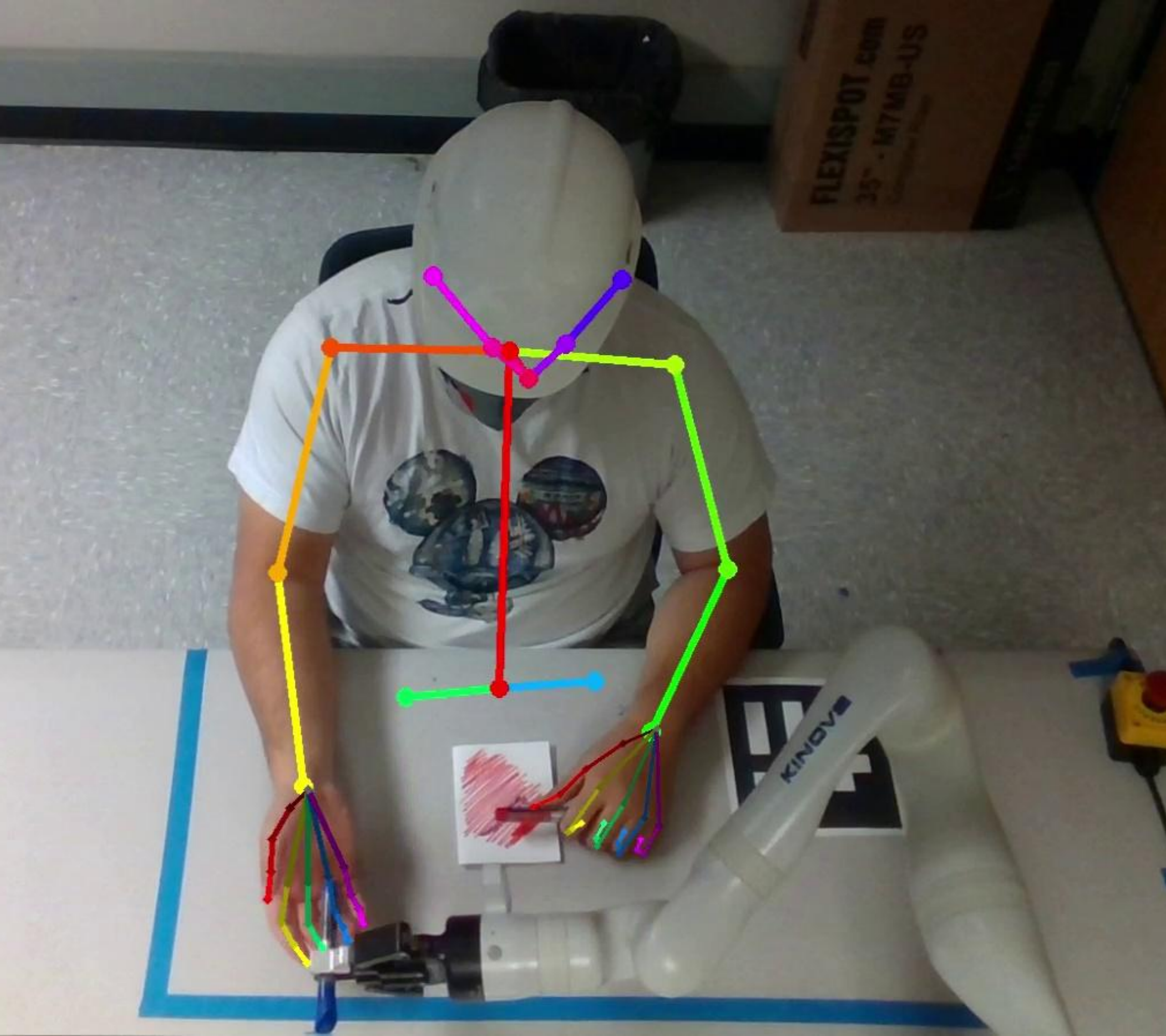}\label{fig:light_bright_deliver}}\hfill
\subfigure[\footnotesize Dark: handover with a safe distance.]{\includegraphics[width=0.45\linewidth]{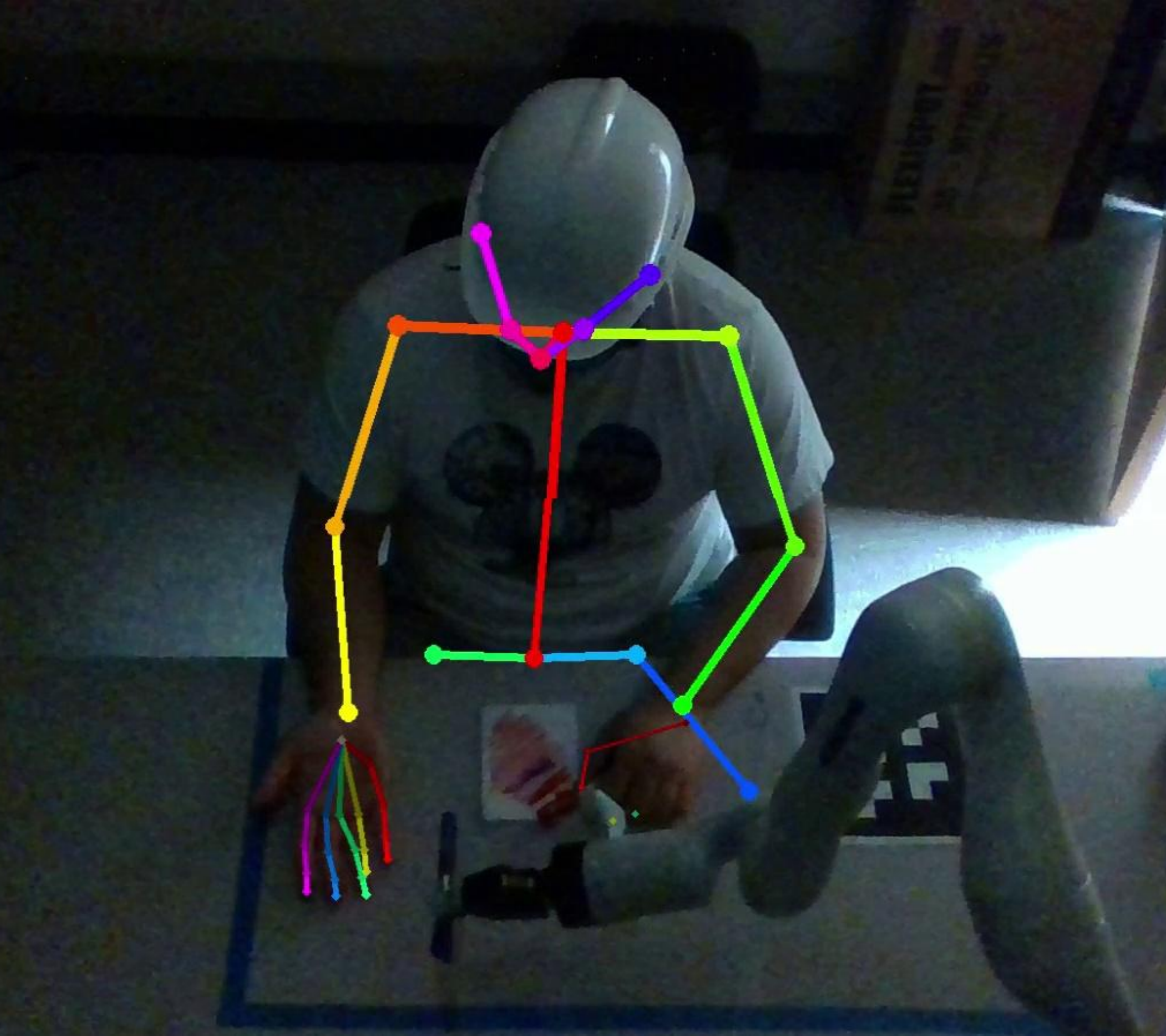}\label{fig:light_dark_deliver}}
\vspace{-10pt}
    \caption{\footnotesize Human-robot handover in different lighting conditions. The skeleton detection is shown in colored line segments. \label{fig:light_deliver}}
\end{figure}

The end-effector goal $x_G$ is determined by perception algorithms, \ie OpenPose \cite{openpose}.
In human-robot handover, we use skeleton detection (\cref{fig:light_deliver}) to locate the delivery position near the human hand.
However, due to the lighting condition (intensity, direction, etc.), the skeleton detection accuracy may vary. 
For example, \cref{fig:light_dark_deliver} contains false detection due to insufficient illumination.
In such cases, the detected delivery position might not always be accurate or safe.
Hence, it is necessary to adjust the goal $x_G$ accordingly.

We observe that the perception uncertainty varies under different conditions for different platforms. 
In particular, we investigated the perception uncertainty of two systems, Kinect v2+MATLAB and Realsense D435i+OpenPose \cite{openpose}.
\Cref{table:perception} shows the perception uncertainty of the two perception systems under different lighting conditions.
The uncertainty is calculated based on the detected right wrist position in 10 consecutive frames. 
Under good lighting conditions, both perception systems have accurate and stable skeleton detection with sub-centimeter uncertainty.
However, the detection robustness drops significantly for the RealSense system when the environment gets dark, which increases to 1-10 cm.
On the other hand, the performance remains similar for the Kinect system.
This is due to the internal tracking of the Kinect system whereas the RealSense system performs single-frame detection. 
The above observation indicates that the perception uncertainty is a reasonable indicator of human detection quality. And environment lighting condition would indeed influence the quality of the human detection, and thus, affect safety during the interaction.

\begin{table}
\begin{center}
\begin{tabular}{c  |  c  c  c | c c c} 
\toprule
  & \multicolumn{3}{c}{Kinect+MATLAB} & \multicolumn{3}{c}{RealSense+OpenPose}\\
 & X & Y & Z & X & Y & Z\\  
\midrule
\multirow{1}{*}{Bright} 
   & 0.31 & 0.76 & 0.57 & 0.10 & 0.40 & 0.75\\
\midrule
\multirow{1}{*}{Dark} 
   &  0.27 & 0.58 & 0.42 & 0.98 & 6.08 & 10.69 \\
\bottomrule
\end{tabular}
\caption{\footnotesize Comparison of the perception uncertainty (\ie standard deviation) of the skeleton keypoint detection using different perception systems in different lighting conditions. All values are in centimeters (cm).}\label{table:perception}
\vspace{-10pt}
\end{center}
\end{table}

We handle the perception uncertainty within solely $g(\cdot)$ by enlarging the distance between goal and human when sensor noise is significant. This is a design specific to this work, since the perception uncertainty can also be handled in other manners such as constraining the robot velocity.
Specifically, given a nominal pose $x_G'=[p_G'; \omega_G']$ from the perception system, where $p_G'$ and $\omega_G'$ are the translation and rotation components, we push the robot handover pose $x_G'$ towards a safe direction by a proportional amount to the standard deviations of human skeleton keypoint detection.
The adapted handover pose is calculated as 
\begin{equation}
\label{eq:perception}
    \begin{split}
        x_G=g(E)=\begin{bmatrix}
        p_G'+\begin{bmatrix}
        \lambda_x\sigma_x\\
        \lambda_y\sigma_y\\
        \lambda_z\sigma_z\\
        \end{bmatrix}
        \cdot u_\mathrm{safe}\\
        \omega_G'
        \end{bmatrix},
    \end{split}
\end{equation}
where $\lambda_{x/y/z}$ are tunable scalers, $\sigma_{x/y/z}$ are standard deviations of detections in X/Y/Z axis, and $u_\mathrm{safe}$ is a unit vector pointing away from the human.
Both $\lambda_{x/y/z}$ and $u_\mathrm{safe}$ should be designed according to actual tasks.
For instance, in a real handover task (\cref{fig:light_deliver}) where the robot delivers a pen to the human, the vector $u_\mathrm{safe}$ can be designed as a unit vector in the direction of $[0;1;1]$ (assuming x-axis is pointing to left and y-axis is pointing downward in \cref{fig:light_deliver}), which offsets the delivery pose away from the human.

\subsection{Adaptation to Different Human Subjects}

\begin{figure}
\subfigure[]{\includegraphics[width=0.24\linewidth]{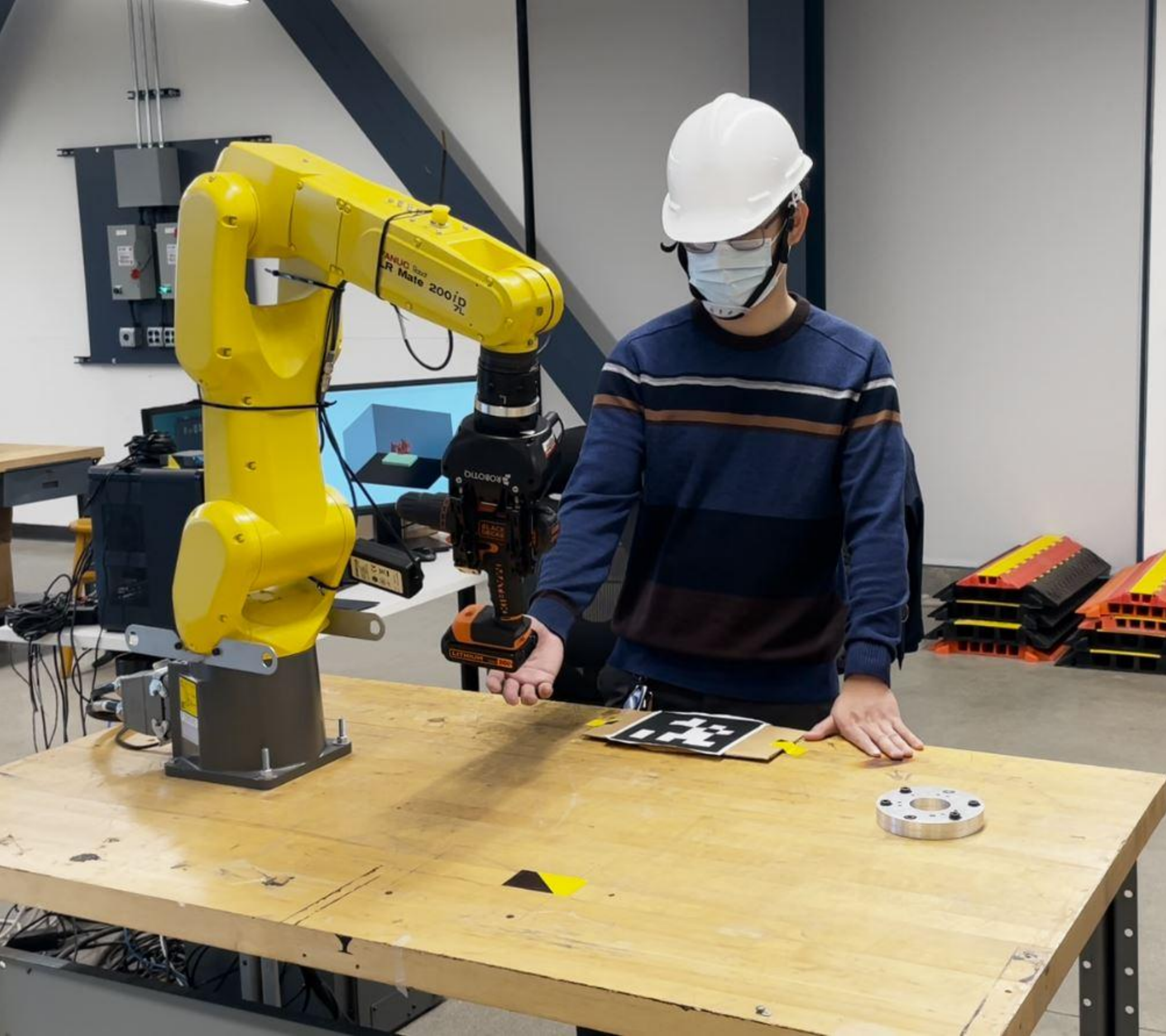}\label{fig:human1}}\hfill
\subfigure[]{\includegraphics[width=0.24\linewidth]{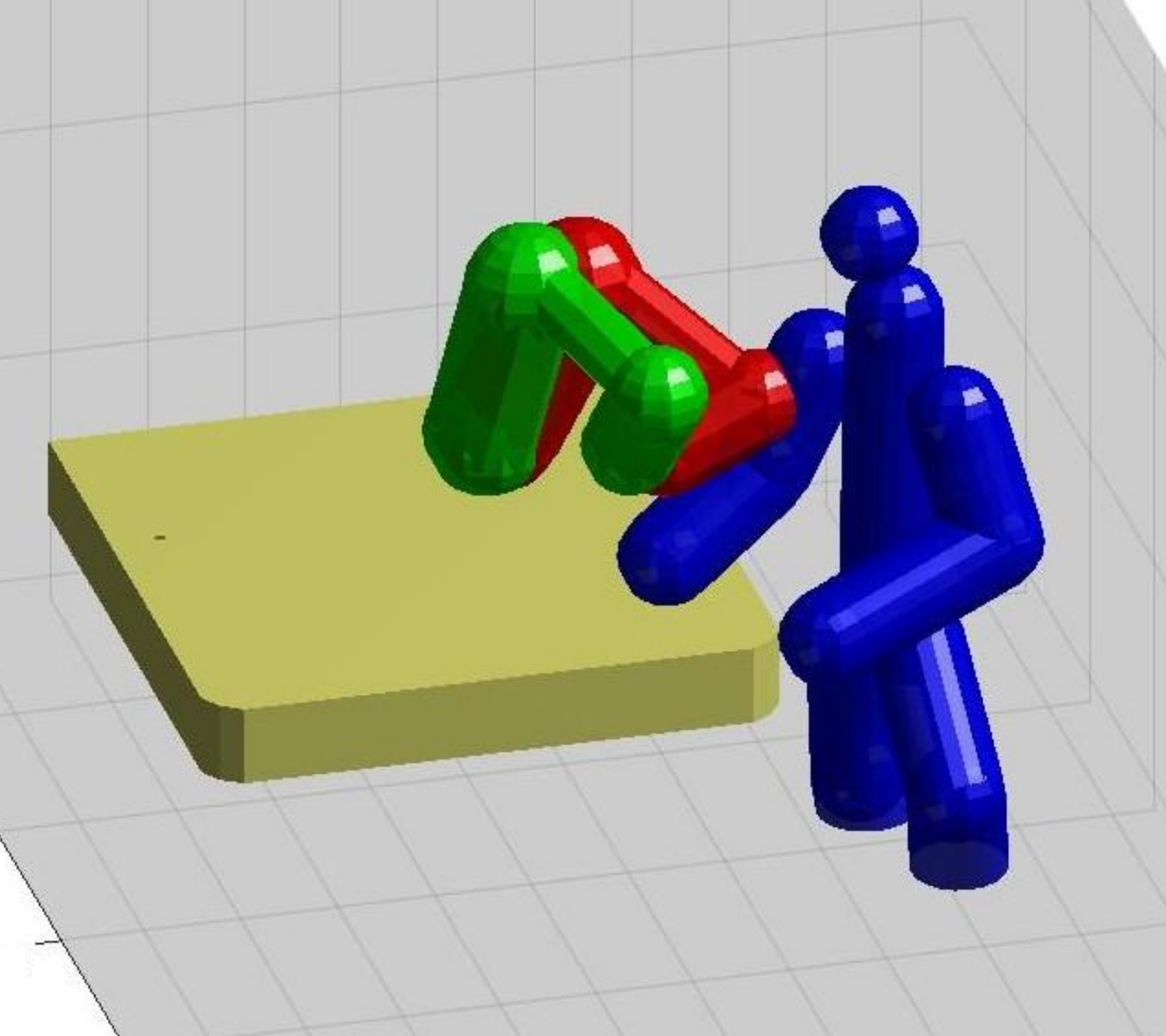}\label{fig:optimalik1}}\hfill
\subfigure[]{\includegraphics[width=0.24\linewidth]{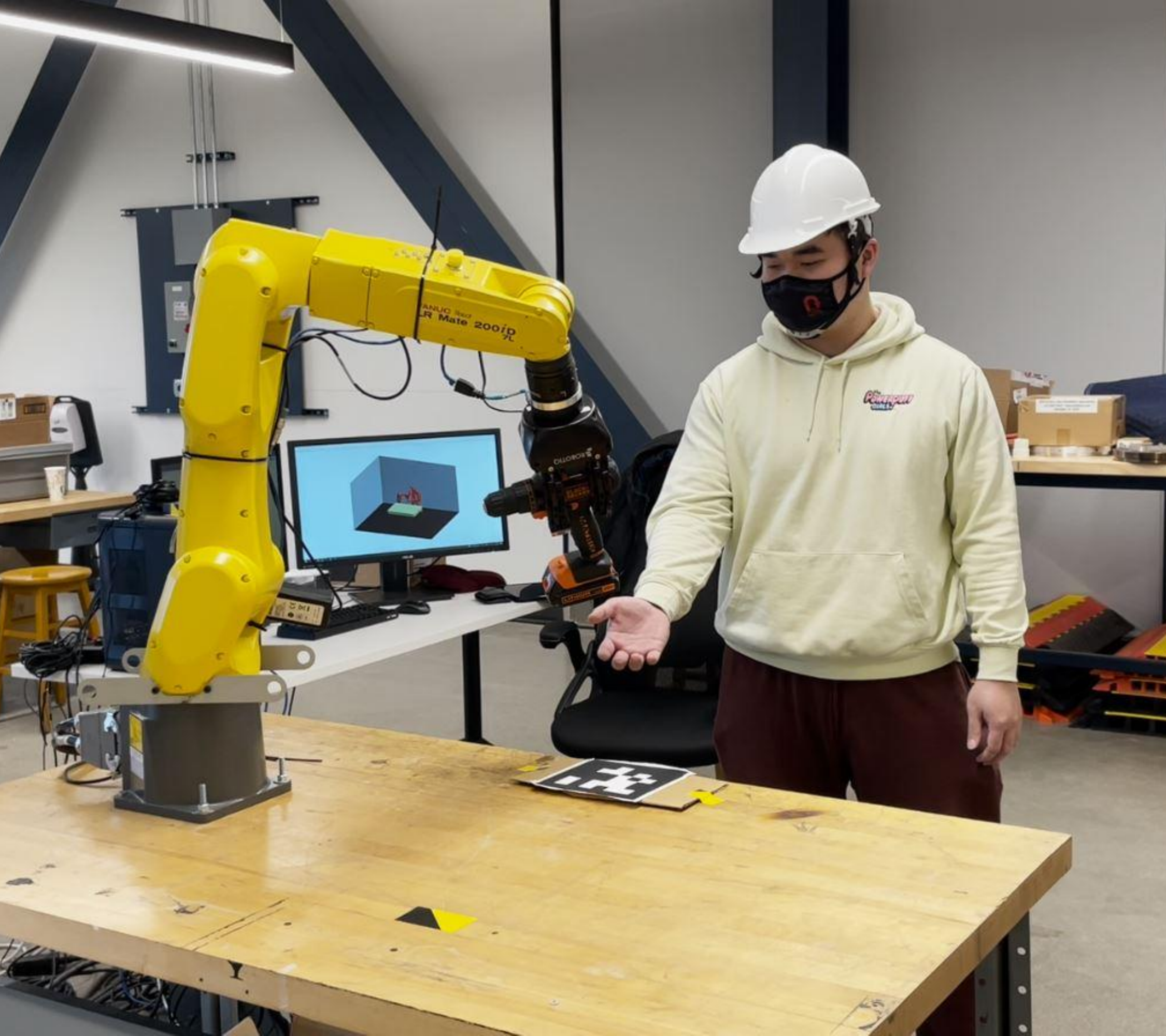}\label{fig:human2}}\hfill
\subfigure[]{\includegraphics[width=0.24\linewidth]{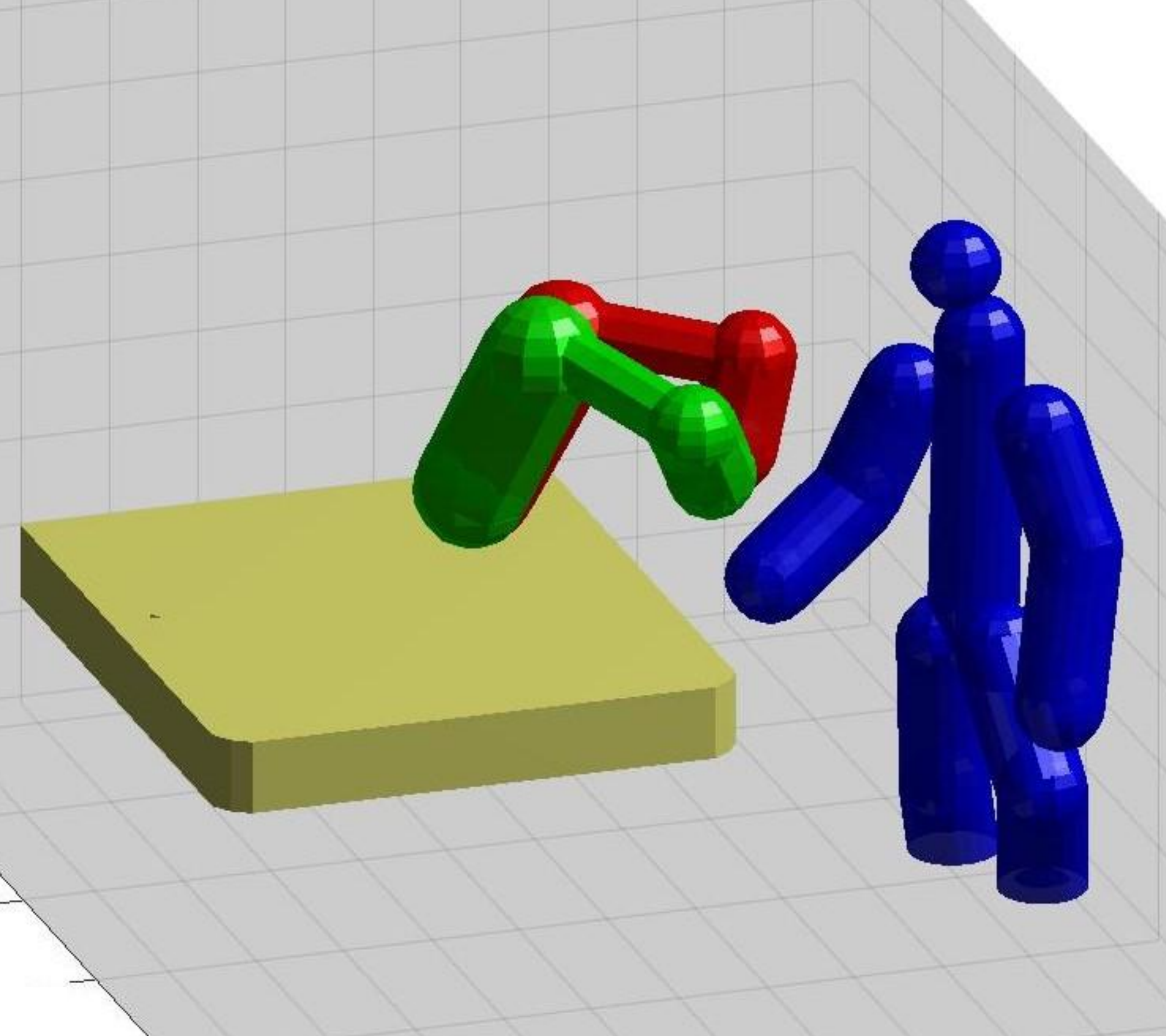}\label{fig:optimalik2}}
\vspace{-10pt}
    \caption{\footnotesize Human-robot handover with different human users. (a) and (c) display the robot delivering the power drill to different humans. (b) and (d) correspond to the visualizations of the adapted delivery after optimization.
    Blue: capsule representation of the human body. Red: initial robot handover configuration. Green: optimized robot handover configuration according to the human pose. \label{fig:humans}}
\end{figure}

Different humans are likely to interact with robots in different ways.
For example, \cref{fig:humans} shows the robot delivering a power drill to the right hands of two different humans. In \cref{fig:human1}, the user stands close to the robot with his left arm relaxing on the table, while the human in \cref{fig:human2} is more conservative and stands further away from the robot. 
Therefore, the robot has more freedom of movement when interacting with the user in \cref{fig:human2} and should be more conservative when interacting with the human user in \cref{fig:human1} to ensure safety.

We adapt the robot delivery pose according to human behavior to ensure safety. 
Specifically, given an end-effector delivery goal $x_G=[p_G;\omega_G]$, we desire the robot end-effector to reach the position $p_G$ while keeping all its links away from the human body to ensure safety. 
We solve this problem by perturbing the nominal delivery joint goal $q_G=IK(x_G)$, where $IK(\cdot)$ is the inverse kinematics mapping \cite{Welman1993INVERSEKA}.
The problem can be formulated as 
\begin{equation}\label{eq:user_adapt}
    \begin{split}
        \argmin_{q_g}&\underbrace{-d(q_g, E)+\lambda_{\omega}||e_{\omega}||}_{V},\\ 
        \st & p_G = \begin{bmatrix}
        I_3 & 0
        \end{bmatrix}\cdot FK(q_g)=\begin{bmatrix}
        I_3 & 0
        \end{bmatrix}\cdot FK(q_G)\\
        & e_{\omega} = \begin{bmatrix}
        0 & I_3
        \end{bmatrix}\cdot FK(q_g)
        -\begin{bmatrix}
        0 & I_3
        \end{bmatrix}\cdot FK(q_G),
    \end{split}
\end{equation}
where $FK(\cdot)$ is the robot forward kinematics mapping and $I_3\in \mathbb{R}^{3\times3}$ is an identity matrix.
The minimizing objective $V$ has two parts. The first part $-d(q_g, E)$ maximizes the distance between the robot links and the human. 
The second part $\lambda_{\omega}||e_{\omega}||$, where $\lambda_{\omega}$ is a scaling weight, regulates the orientation error $e_{\omega}$.
Note that the translation is enforced as a hard constraint whereas the orientation is a soft constraint since we allow deviation in the orientation $\omega_G$ but keep the delivery pose $p_G$ invariant.
The user adaptation in \cref{eq:user_adapt} can be solved numerically as shown in \cref{alg:user}.

\begin{algorithm}\small
    \caption{User Adaptation \label{alg:user}} 
	\begin{algorithmic}[1]
	    \State Initialize: $V=\infty$, $q_G=IK(x_G)$, $q_g=q_G$, $q=q_G$.
	    \State Output: $q_g$.
	    \While{not reach maximum iterations}
		\State Find $\Delta q\in \mathbf{N}(J(q))$. \Comment{Search in the null space direction.}
		\State $q \longleftarrow q+\alpha \Delta q$. \Comment{Perturb the joint goal.}
		\State $q \longleftarrow ICOP(q, p_G)$. \Comment{Ensure Cartesian position tracking.} 
		\State $V' \longleftarrow -d(q, E) + \lambda_{\omega}||e_{\omega}||$. \Comment{Compute objective cost in \cref{eq:user_adapt}.}
		\If{$V'<V$} \Comment{Update solution.}
		\State $V\longleftarrow V'$.
		\State $q_g\longleftarrow q$.
		\EndIf
		\EndWhile
	\end{algorithmic} 
\end{algorithm}
$\mathbf{N}(\cdot)$ denotes the null space and $J(\cdot)$ calculates the robot jacobian given the configuration $q$.
To enforce the translation equality constraint, we invoke ICOP algorithm \cite{cfs} after the perturbation on line 6. 
This is necessary because the null space search direction $\Delta q$ will only preserve the robot end-effector position instantaneously at joint state $q$, and is no longer valid once $q$ is perturbed, \ie $q+\Delta q$. 
To preserve position tracking in the iterative optimization, $q$ should be corrected after each perturbation.
After solving \cref{alg:user}, we get an adapted joint goal $q_g$ which still tracks nominal delivery position $p_G$, but with a larger distance between the human and the robot to ensure interactive safety.

\subsection{Adaptation to Robot Hardware}

\begin{figure}
    \centering
    \includegraphics[width=0.85\linewidth]{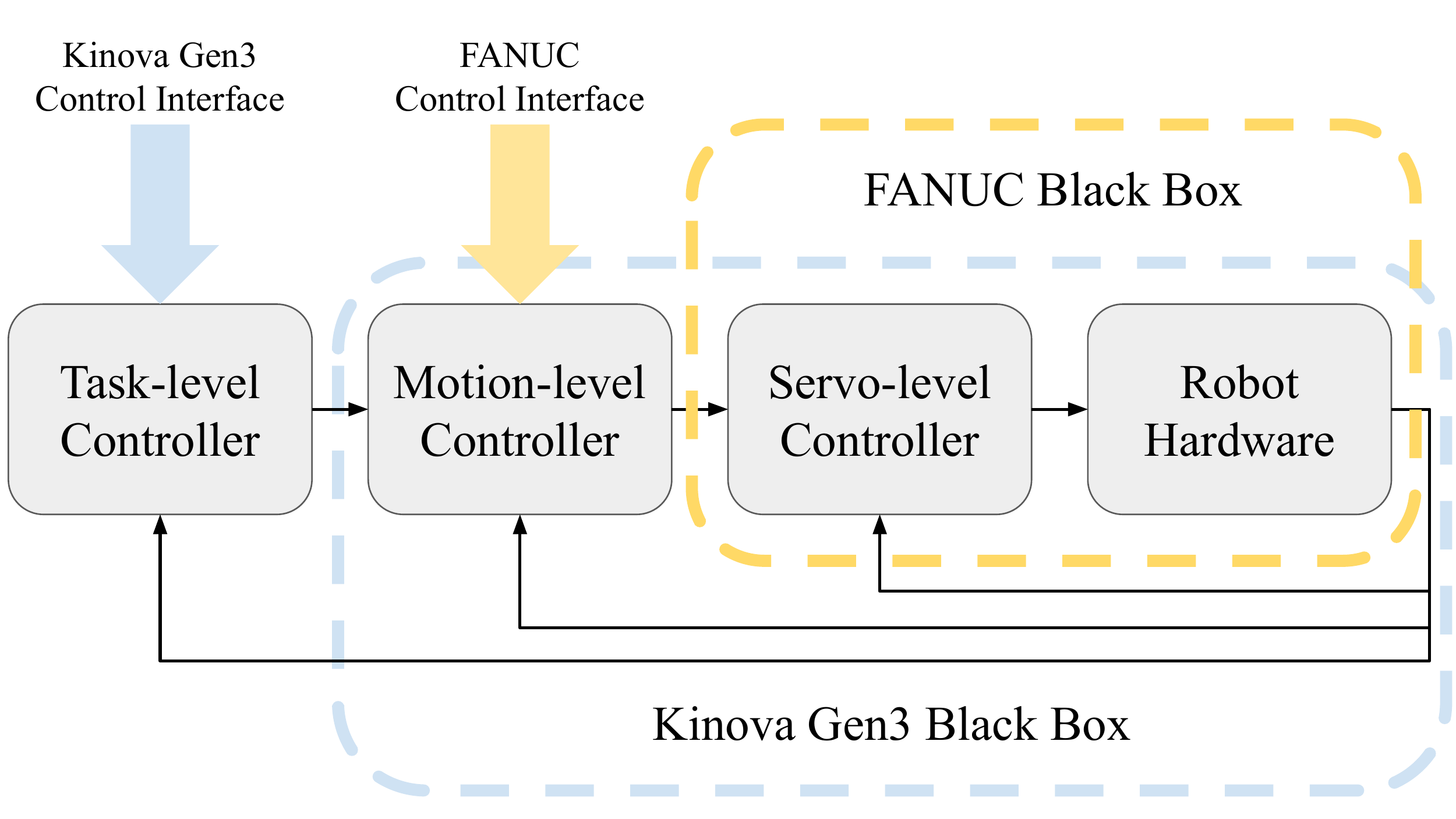}
     \vspace{-10pt}
    \caption{\footnotesize Control schemes of the FANUC and Kinova robots.}
    \label{fig:control_scheme}
\end{figure}

The need for adaptation to different hardware platforms mainly stems from different robot control interfaces. 
We view the control in robot tasks to be the cascade of three levels: (1) task-level, which computes sparse tracking references to fulfill high-level tasks, (2) motion-level, which generates dense tracking references to accurately specify robot trajectories, and (3) servo-level, which computes the motor torques needed to track desired robot motions.
\Cref{fig:control_scheme} shows a comparison of FANUC and Kinova robots in terms of the available control interfaces. 
The principle of adaptation to hardware is to share modules that are outside the level of control provided by both robots' control interfaces and develop hardware-specific modules to fill any gap.

As shown in \cref{fig:control_scheme}, the FANUC robot operates on motion level while the Kinova robot operates on task level. 
Hence, modules in task-level control can be shared across two robots, such as process control of the handover pipeline, robot kinematics, and adaptation to lighting conditions and human users.
However, other modules in inner control levels cannot be shared. 
For both robots, we assume that the shared task-level controller has already generated a single-point joint tracking reference $q_g$ in \cref{eq:prob}, and focus on how we enable each robot to track that reference (\ie tracking controller) while guaranteeing human safety (\ie safety controller).

\subsubsection{3.3.1 Tracking Controller}\label{sec:tracking_controller}

We use a jerk-bounded position controller (JPC) \cite{jpc} as the motion-level tracking controller for the FANUC robot. 
Given a tracking reference $q_g$, the JPC calculates a jerk control sequence $j=JPC(q_g)$. 
The position control sequence is calculated by integrating the jerk sequence and sent for execution.

Since the Kinova robot operates on a task-level control interface, the Kinova controller does not need to pre-plan the whole trajectory towards the reference. Hence, we implement the Kinova controller as a feedback controller with a low update rate ($<$30Hz). Specifically, at each time $k$, given a joint tracking reference $q_k$, the Kinova controller uses a PD controller to generate a desired joint acceleration $u$ as the nominal tracking control.

\subsubsection{3.3.2 Safety Controller}\label{sec:safe_control}

Since the FANUC robot interface requires bounded jerk, we incorporate the jerk-based safe set algorithm (JSSA) \cite{jssa} to safeguard the control input to the robot. 
Given the nominal control $j_k$, the JSSA generates a safe jerk control command as $j^s_k=JSSA(j_k,q_k,E_k)$, where $q_k$ and $E_k$ denote the real-time robot and environment configurations.
The position control is calculated by integrating the safe jerk control $j^s_k$.

To modify the nominal Kinova acceleration control $u_k$ for safety, we incorporate the acceleration-based SSA (SSA) \cite{ssa} in the Kinova controller.

\begin{figure}
\subfigure[]{\includegraphics[width=0.3\linewidth]{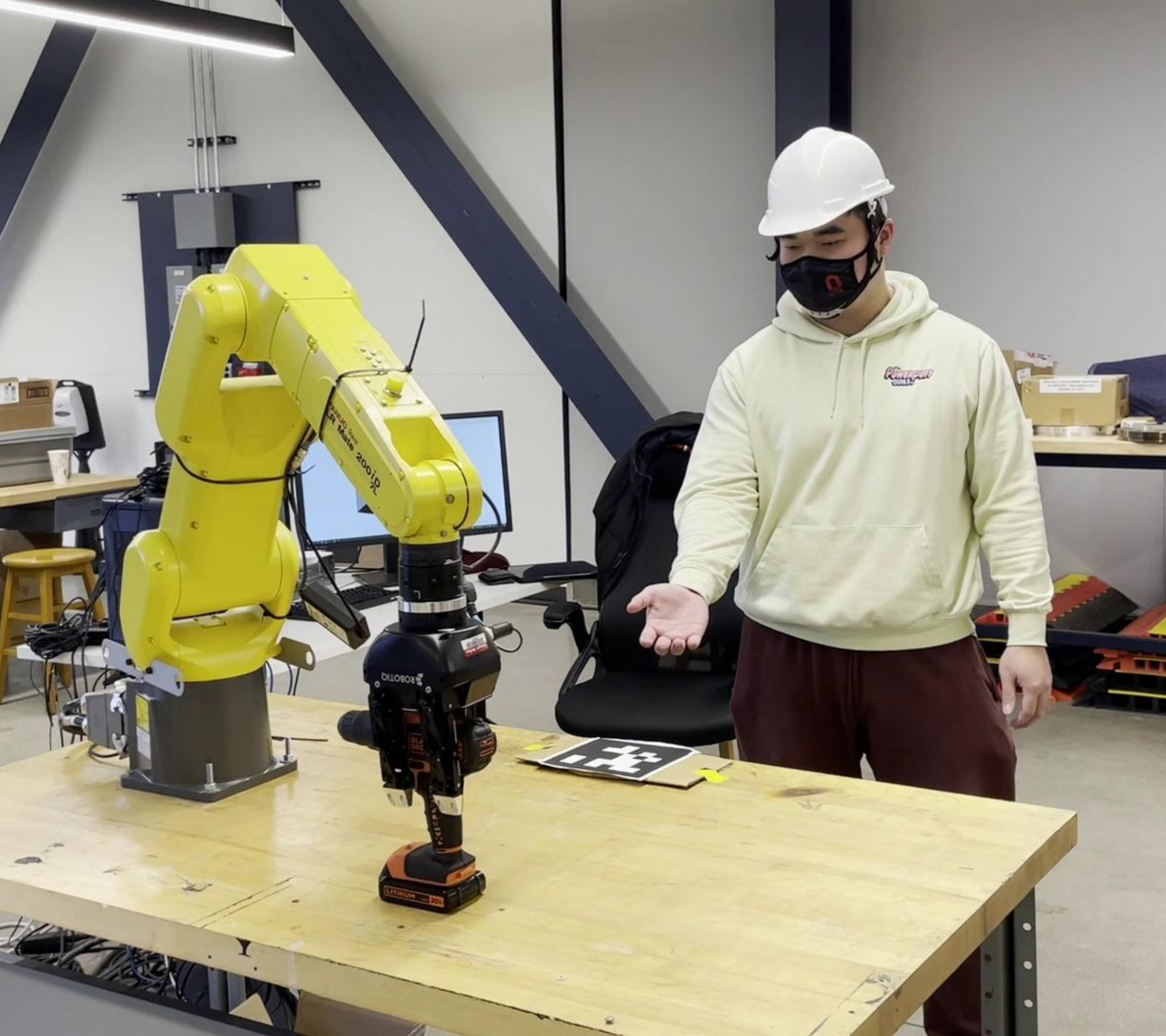}\label{fig:fanuc2}}\hfill
\subfigure[]{\includegraphics[width=0.3\linewidth]{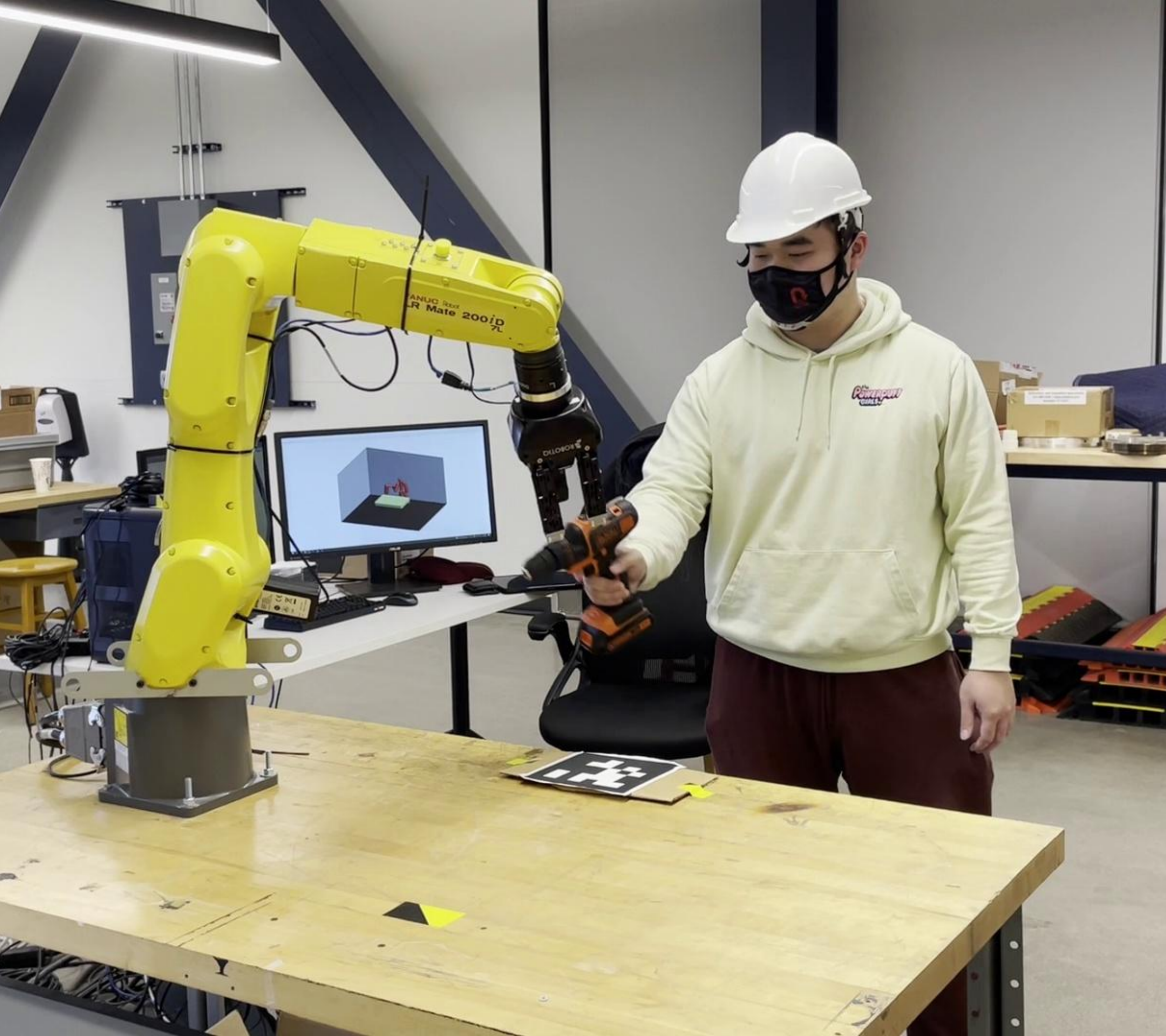}\label{fig:fanuc3}}\hfill
\subfigure[]{\includegraphics[width=0.3\linewidth]{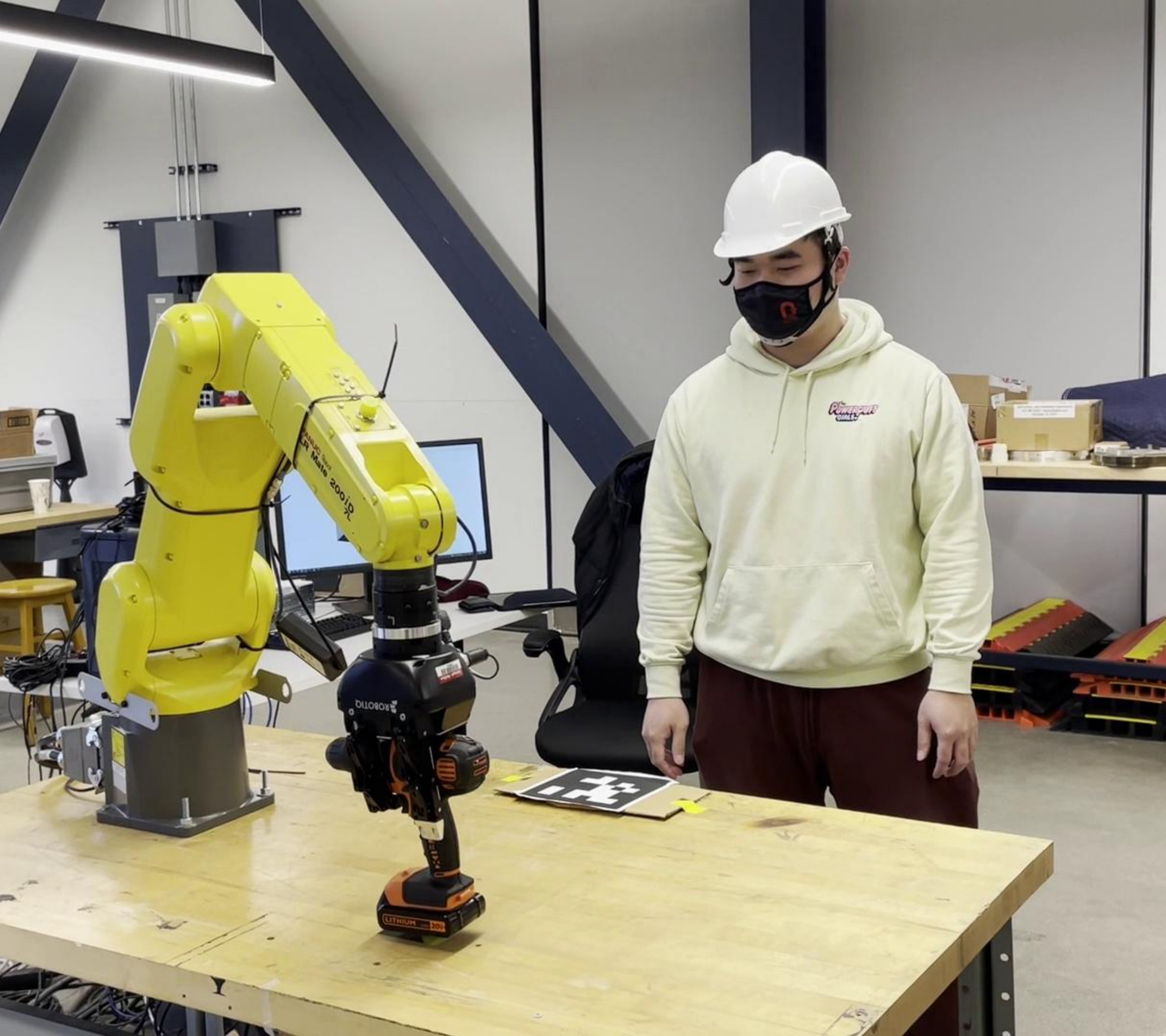}\label{fig:fanuc5}}\\
\subfigure[]{\includegraphics[width=0.3\linewidth]{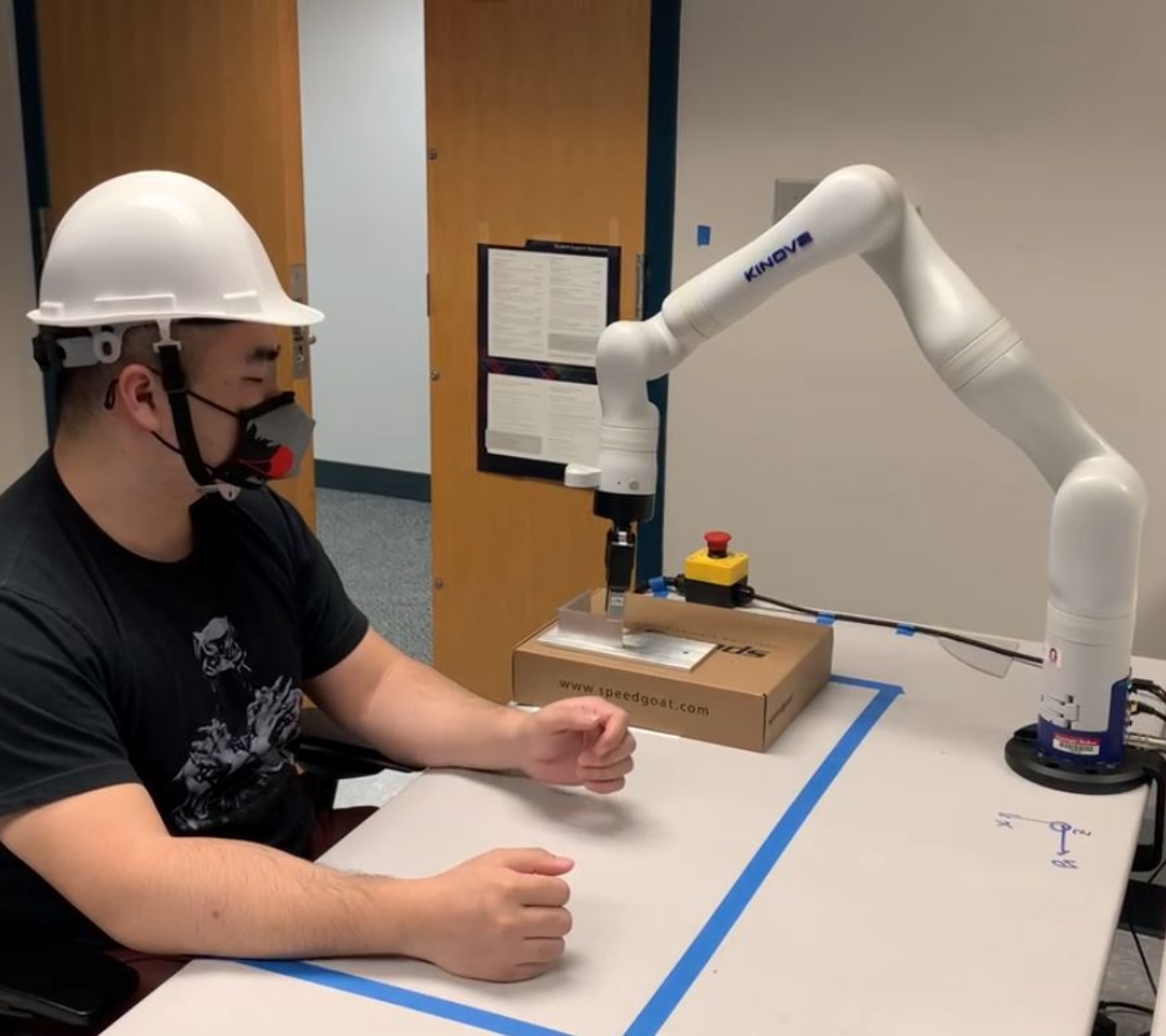}\label{fig:kinova1}}\hfill
\subfigure[]{\includegraphics[width=0.3\linewidth]{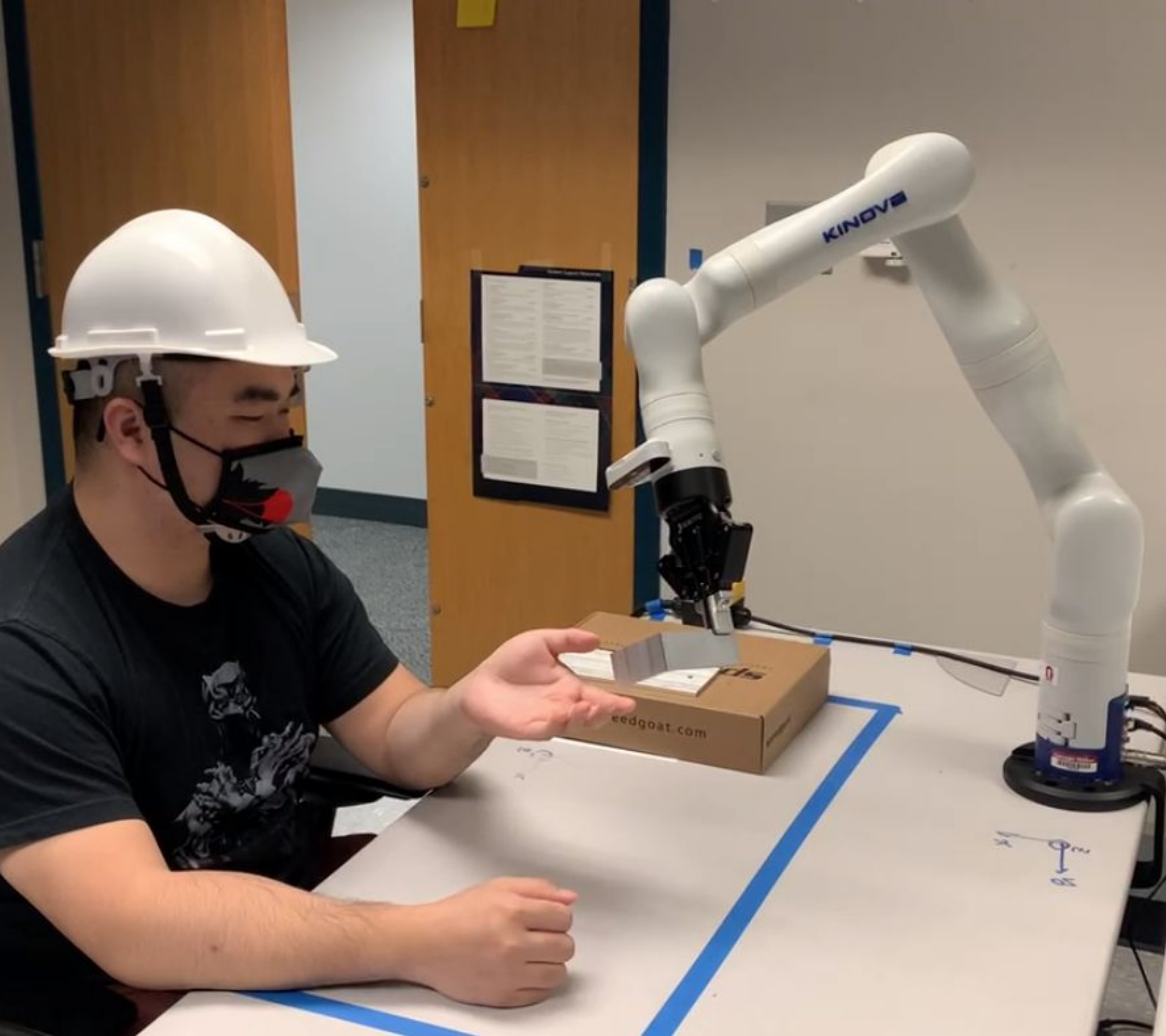}\label{fig:kinova2}}\hfill
\subfigure[]{\includegraphics[width=0.3\linewidth]{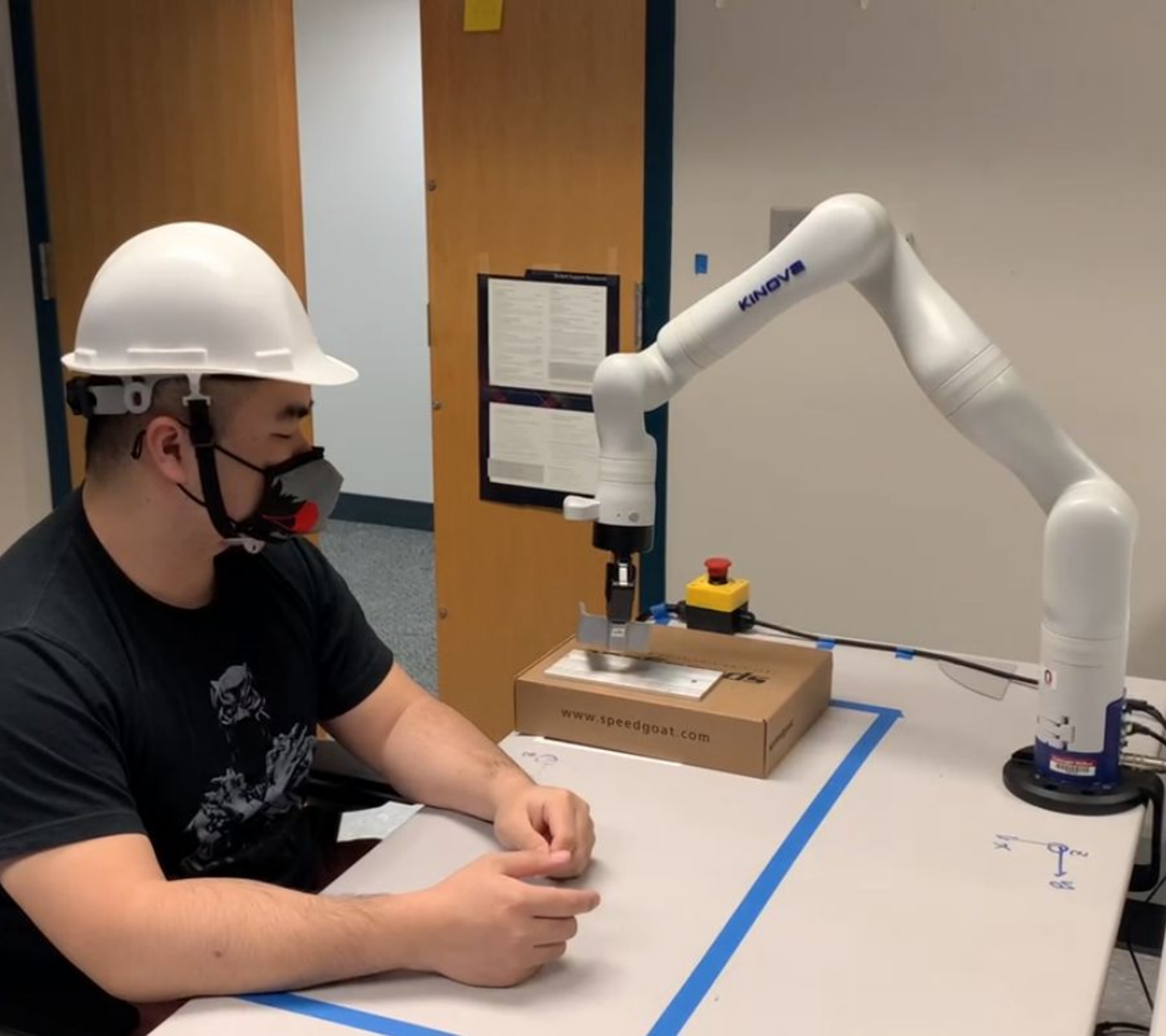}\label{fig:kinova5}}
\vspace{-10pt}
    \caption{\footnotesize Human-robot handover with FANUC LR Mate 200id/7L and Kinova Gen3. \label{fig:fanuc_handover}}
\end{figure}

\section{Experiment Results}\label{sec:exp}
In this section, we demonstrate the task-agnostic adaptation for safe human-robot handover in different lighting conditions, with different humans, and with different robots.
Due to the high-frequency communication requirement of the SG (discussed in \cref{sec:so}), the tracking and safe controllers for the FANUC robot are deployed to the SG. The remaining modules are running on PC.

\subsection{Safe Handover with Different Lighting Conditions}

We first demonstrate the robot handover in different lighting conditions.
In this experiment, we have the same robot (\ie Kinova Gen3) and the same human performing the handover task.

\Cref{fig:light_deliver} shows the robot delivering poses with bright and dark lighting conditions.
When having a good lighting condition (\cref{fig:light_bright_deliver}), the perception system has stable and accurate human skeleton pose detection.
Therefore, the system has high confidence and delivers the object straight to the human hand.
On the other hand, when having a dark environment (\cref{fig:light_dark_deliver}), the skeleton detection becomes unstable, leading to large uncertainty.
The uncertainty makes the robot behave more conservatively and deliver the pen slightly away from the human in order to ensure no collision with the human.

\subsection{Safe Handover with Different Humans}

We also demonstrate the robot handover with different human subjects.
In this experiment, we have the same robot (\ie FANUC LR Mate 200id/7L) and the same lighting condition when performing the handover task.

The visualizations of the handover tasks with the two humans are shown in \cref{fig:optimalik1} and \cref{fig:optimalik2}, which correspond to the situations in \cref{fig:human1} and \cref{fig:human2}.
The blue capsules represent the human body, which are captured using a Kinect V2.
The delivery location is determined by the right hand of the human.
Given the delivery position, the red capsules indicate the initial reference robot delivery configuration, which is solved using inverse kinematics \cite{Welman1993INVERSEKA}. 
The green capsules represent the optimized robot delivery configuration in order to adapt to different human behavior.
The human in \cref{fig:human1} stands close to the robot with his left hand resting on the table. 
The reference delivery configuration collides with the human body as shown in \cref{fig:optimalik1}.
Therefore, the adaptation adjusts the robot intermediate links while keeping the end-effector at the same delivery location, and makes the handover safe and comfortable for the human user.
On the other hand, the human in \cref{fig:human2} stands further away from the robot.
Thus, the reference robot configuration is safe as shown in \cref{fig:optimalik2}. 
In this case, the adaptation still toggles the intermediate links to stay away from the human and makes the interaction comfortable for the human.
In our experiments, solving optimal robot delivery pose typically takes less than $0.5s$ by running \cref{alg:user}.

\subsection{Safe Handover with Different Robots}

\begin{figure}
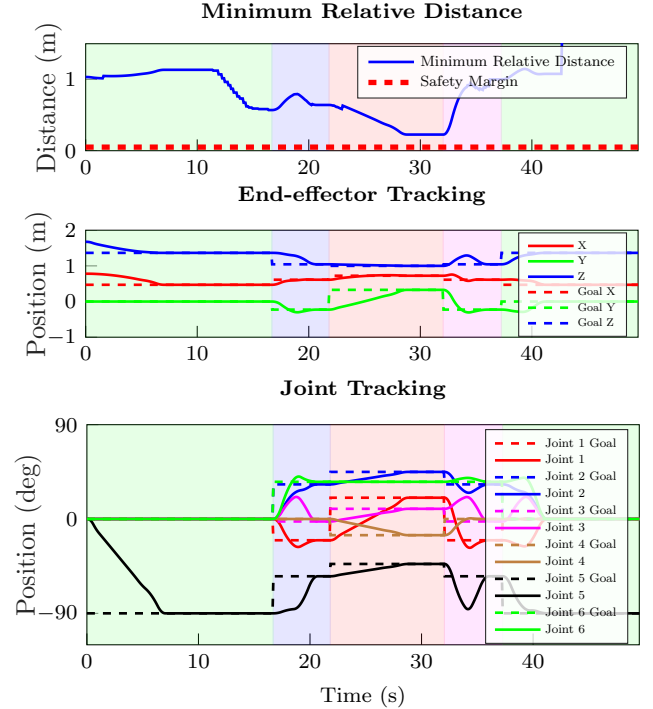

\centering
\input{plot/fanuc}
\input{plot/fanuc_joint}
\vspace{-10pt}
\caption{\footnotesize Human-robot handover profiles with FANUC LR Mate 200id/7L.}
    \label{fig:fanuc_handover_profile}
\end{figure}

\begin{figure}
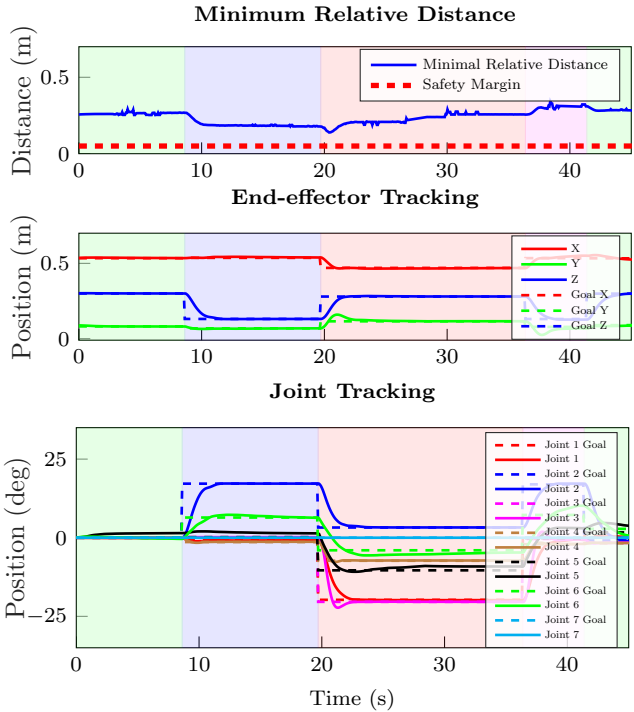

\centering
\input{plot/kinova}
\input{plot/kinova_joint}
\vspace{-10pt}
\caption{\footnotesize Human-robot handover profiles with Kinova Gen3.}
    \label{fig:kinova_handover_profile}
\end{figure}

In this section, we demonstrate the human-robot handover with different robots (\ie Kinova Gen3 and FANUC LR Mate 200id/7L).
We have the same human user and similar bright lighting when performing the handover task.

\Cref{fig:fanuc_handover} breakdowns the complete handover process with the two robots.
The FANUC robot performs a tool handover, in which the robot delivers a power drill to the human who stands in front of the table.
The human raises his palm when he wants the tool and the robot recognizes the human intention and grabs the power drill (\cref{fig:fanuc2}).
The robot then localizes the human palm position, optimizes its delivery configuration, and hands the power drill to the human (\cref{fig:fanuc3}).
After the human finishes using the power drill, the robot then places the power drill back in its original position for tool management (\cref{fig:fanuc5}).
And similarly, the Kinova robot performs a workpiece handover, in which the robot delivers a metal piece to the human who sits in front of the table.
The robot picks up the metal piece (\cref{fig:kinova1}) and delivers it to the human with an appropriate configuration (\cref{fig:kinova2}).
The robot also returns the metal piece when the human finishes inspecting the workpiece (\cref{fig:kinova5}).

\Cref{fig:fanuc_handover_profile,fig:kinova_handover_profile} demonstrate the detailed profiles of the robot handover with the FANUC and Kinova robots. The green area indicates the period that the robot has no task to execute, meaning that it is idle and waiting for a human command.
The blue area indicates that the robot recognizes the human intention and grabs the desired object on the table.
The red area indicates the period that the robot is delivering the object, and the purple area is when the robot is returning the object.
We observe that even though we have two completely different robot platforms, the task-agnostic adaptation enables easy transferability and makes the two robots have very similar behavior in terms of safe handover.
As shown in the top plots in \cref{fig:fanuc_handover_profile} and \cref{fig:kinova_handover_profile}, safety is ensured since the minimum distance between the robot and the human is consistently greater than the safety margin.
From the middle figures, we can see the robot is properly executing the task as the end-effector tracks the goals for all subtasks.
The bottom plots show the tracking performance for each joint.
The robots are properly tracking the goals safely to finish the handover task.

\section{Conclusion}\label{sec:conclusion}

In this paper, we present a task-agnostic adaptable controller that can accommodate varying objectives, constraints, and executions in general HRI tasks. We present our controller within the context of human-robot handover, and show how it can adjust safety margin in varying lighting conditions, optimize handover movements for user safety and comfort, and adapt to different robot platforms. We verify our approach in handover tasks under distinctive lighting conditions, user interaction styles, and robot control interfaces. For future work, we aim to apply our controller to more complicated tasks such as collaborative assembly, where the adaption would also need to consider task-specific information such as the assembly process.

\bibliography{ifacconf}          

\begin{thebibliography}{13}
\providecommand{\natexlab}[1]{#1}
\providecommand{\url}[1]{\texttt{#1}}
\providecommand{\urlprefix}{URL }
\expandafter\ifx\csname urlstyle\endcsname\relax
  \providecommand{\doi}[1]{doi:\discretionary{}{}{}#1}\else
  \providecommand{\doi}{doi:\discretionary{}{}{}\begingroup
  \urlstyle{rm}\Url}\fi

\bibitem[{{Cao} et~al.(2019){Cao}, {Hidalgo Martinez}, {Simon}, {Wei}, and
  {Sheikh}}]{openpose}
{Cao}, Z., {Hidalgo Martinez}, G., {Simon}, T., {Wei}, S., and {Sheikh}, Y.A.
  (2019).
\newblock Openpose: Realtime multi-person 2d pose estimation using part
  affinity fields.
\newblock \emph{IEEE Transactions on Pattern Analysis and Machine
  Intelligence}.

\bibitem[{Christensen et~al.(2021)Christensen, Amato, Yanco, Matari{\'c},
  Choset, Drobnis, Goldberg, Grizzle, Hager, Hollerbach, Hutchinson, Krovi,
  Lee, Smart, Trinkle, and Sukhatme}]{Christensen2021ARF}
Christensen, H.I., Amato, N.M., Yanco, H.A., Matari{\'c}, M.J., Choset, H.,
  Drobnis, A.W., Goldberg, K., Grizzle, J.W., Hager, G., Hollerbach, J.M.,
  Hutchinson, S., Krovi, V.N., Lee, D., Smart, B., Trinkle, J.C., and Sukhatme,
  G.S. (2021).
\newblock A roadmap for us robotics - from internet to robotics 2020 edition.
\newblock \emph{Found. Trends Robotics}, 8, 307--424.

\bibitem[{Gopinath et~al.(2017)Gopinath, Ore, and Johansen}]{GOPINATH2017430}
Gopinath, V., Ore, F., and Johansen, K. (2017).
\newblock Safe assembly cell layout through risk assessment – an application
  with hand guided industrial robot.
\newblock \emph{Procedia CIRP}, 63, 430--435.

\bibitem[{Krüger et~al.(2009)Krüger, Lien, and Verl}]{KRUGER2009628}
Krüger, J., Lien, T., and Verl, A. (2009).
\newblock Cooperation of human and machines in assembly lines.
\newblock \emph{CIRP Annals}, 58(2), 628--646.

\bibitem[{Liu et~al.(2018{\natexlab{a}})Liu, Lin, and Tomizuka}]{cfs}
Liu, C., Lin, C.Y., and Tomizuka, M. (2018{\natexlab{a}}).
\newblock The convex feasible set algorithm for real time optimization in
  motion planning.
\newblock \emph{SIAM Journal on Control and Optimization}, 56(4), 2712--2733.

\bibitem[{Liu et~al.(2018{\natexlab{b}})Liu, Tang, Lin, Cheng, and
  Tomizuka}]{serocs}
Liu, C., Tang, T., Lin, H., Cheng, Y., and Tomizuka, M. (2018{\natexlab{b}}).
\newblock Serocs: Safe and efficient robot collaborative systems for next
  generation intelligent industrial co-robots.
\newblock \emph{CoRR}, abs/1809.08215.

\bibitem[{Liu and Tomizuka(2014)}]{ssa}
Liu, C. and Tomizuka, M. (2014).
\newblock Control in a safe set: Addressing safety in human-robot interactions.
\newblock volume~3.

\bibitem[{Liu and Tomizuka(2016)}]{7487476}
Liu, C. and Tomizuka, M. (2016).
\newblock Algorithmic safety measures for intelligent industrial co-robots.
\newblock In \emph{2016 IEEE International Conference on Robotics and
  Automation (ICRA)}, 3095--3102.

\bibitem[{Liu et~al.(2022{\natexlab{a}})Liu, Chen, and Liu}]{jssa}
Liu, R., Chen, R., and Liu, C. (2022{\natexlab{a}}).
\newblock Safe interactive industrial robots using jerk-based safe set
  algorithm.

\bibitem[{Liu et~al.(2022{\natexlab{b}})Liu, Chen, Sun, Zhao, and Liu}]{jpc}
Liu, R., Chen, R., Sun, Y., Zhao, Y., and Liu, C. (2022{\natexlab{b}}).
\newblock Jerk-bounded position controller with real-time task modification for
  interactive industrial robots.
\newblock TechRxiv.

\bibitem[{Welman(1993)}]{Welman1993INVERSEKA}
Welman, C. (1993).
\newblock \emph{Inverse Kinematics and Geometric Constraints for Articulated
  Figure Manipulation [microform]}.
\newblock Canadian theses on microfiche. Thesis (M.Sc.)--Simon Fraser
  University.

\bibitem[{Zhang et~al.(2014)Zhang, Li, Boca, Newkirk, Zhang, Fuhlbrigge, Feng,
  and Hunt}]{6840175}
Zhang, G.Q., Li, X., Boca, R., Newkirk, J., Zhang, B., Fuhlbrigge, T.A., Feng,
  H.K., and Hunt, N.J. (2014).
\newblock Use of industrial robots in additive manufacturing - a survey and
  feasibility study.
\newblock 1--6.

\bibitem[{Zhao et~al.(2020)Zhao, He, Wen, and Liu}]{dscc_cfs}
Zhao, W.Y., He, S., Wen, C., and Liu, C. (2020).
\newblock Contact-rich trajectory generation in confined environments using
  iterative convex optimization.
\newblock Dynamic Systems and Control Conference.

\end{thebibliography}
\end{document}